\pdfoutput=1

\documentclass[11pt]{article}

\usepackage{acl}

\usepackage{times}
\usepackage{latexsym}

\usepackage[T1]{fontenc}

\usepackage[utf8]{inputenc}

\usepackage{microtype}

\usepackage{times}
\usepackage{latexsym}
\usepackage{float}
\usepackage{graphicx}
\usepackage{booktabs}
\usepackage{amsmath,amsfonts,amssymb,amsthm}
\usepackage{bm}
\usepackage[ruled,noend]{algorithm2e}
\usepackage{enumitem}
\usepackage{setspace}
\usepackage{tcolorbox}
\usepackage{soul}
\usepackage{xspace}

\newcommand{\stitle}[1]{\vspace{0.5em}\noindent{\bf #1}}

\usepackage{cleveref}
\crefformat{section}{\S#2#1#3}
\crefformat{subsection}{\S#2#1#3}
\crefformat{subsubsection}{\S#2#1#3}
\crefrangeformat{section}{\S\S#3#1#4 to~#5#2#6}
\crefmultiformat{section}{\S\S#2#1#3}{ and~#2#1#3}{, #2#1#3}{ and~#2#1#3}
\usepackage{refstyle}
\Crefformat{figure}{#2Figure~#1#3}
\Crefmultiformat{figure}{Figures~#2#1#3}{ and~#2#1#3}{, #2#1#3}{ and~#2#1#3}
\Crefformat{table}{#2Table~#1#3}
\Crefmultiformat{table}{Tables~#2#1#3}{ and~#2#1#3}{, #2#1#3}{ and~#2#1#3}
\Crefformat{appendix}{\S#2#1#3}
\crefformat{algorithm}{Alg.~#2#1#3}
\Crefformat{equation}{Eq.~#2#1#3}
\SetKwInput{KwInput}{Input}
\SetKwInput{KwOutput}{Output}
\SetEndCharOfAlgoLine{}

\newcommand{\eg}{e.g.}

\newcommand{\quasi}{\textsc{QuAsi}\xspace}
\usepackage{siunitx}
\title{Answer Consolidation: Formulation and Benchmarking}

\author{Wenxuan Zhou$^{1}$\thanks{\;\;This work was conducted when the first author was doing an internship at Amazon.}, Qiang Ning$^2$, Heba Elfardy$^2$, Kevin Small$^2$, Muhao Chen$^1$\\
$^1$University of Southern California, $^2$Amazon\\
\texttt{\{zhouwenx,muhaoche\}@usc.edu}\; \texttt{\{qning,helfardy,smakevin\}@amazon.com}}

\begin{document}
\maketitle
\begin{abstract}
Current question answering~(QA) systems primarily consider the single-answer scenario, where each question is assumed to be paired with one correct answer. However, in many real-world QA applications, multiple answer scenarios arise where consolidating answers into a comprehensive and non-redundant set of answers is a more efficient user interface. In this paper, we formulate the problem of {\em answer consolidation}, where answers are partitioned into multiple groups, each representing different aspects of the answer set.
Then, given this partitioning, a comprehensive and non-redundant set of answers can be constructed by picking one answer from each group. To initiate research on answer consolidation, we construct a dataset consisting of 4,699 questions and 24,006 sentences and evaluate multiple models. Despite 
a promising performance achieved by the best-performing supervised models, we still believe this task has room for further improvements.\footnote{{The contributed resources and implementation are available at \url{ https://github.com/amazon-research/question-answer-consolidation}.}}


\end{abstract}

\section{Introduction}

Open-domain question answering~(QA) systems~\cite{Voorhees1999TheTQ} 
aim to answer natural language questions using large collections of reference documents, contributing to real-world applications such as intelligent virtual assistants and search engines.
Current QA systems~\cite{zhu2021retrieving} usually adopt a three-stage pipeline consisting of: (1) a \emph{passage retriever}~\cite{yang-etal-2019-end-end,karpukhin-etal-2020-dense} that selects a small set of passages relevant to the question, (2) a \emph{machine reader} that examines the retrieved passages and extracts~\cite{wang-etal-2017-gated,wang-etal-2019-multi} or abstracts~\cite{lewis2020retrieval} the candidate answers, and (3) an \emph{answer reranker}~\cite{wang2018evidence,kratzwald-etal-2019-rankqa} that fuses features from previous stages to either select one final answer or return the top-ranked answer from the previous stage.

Current QA research~\cite{joshi-etal-2017-triviaqa,kwiatkowski-etal-2019-natural} primarily examines the case where each question is assumed to have a single correct answer.
However, in practice, many questions can have multiple correct answers.
For example, the question ``\emph{Is coffee good for your health?}'' can be answered with respect to different aspects (\eg, ``\emph{coffee can help you with weight loss}'', ``\emph{coffee can cause insomnia and restlessness}''). 
To correctly identify different aspects of answers to the same question while mitigating aspect-level redundancy, it is important to \emph{consolidate} the answers. Answer consolidation is particularly desirable for applications such as intelligent assistants, where responses are desired to be both comprehensive and concise.
Additionally, in scenarios where QA is used for knowledge extraction~\cite{bhutani-etal-2019-open,du-cardie-2020-event} or claim verification~\cite{yin-roth-2018-twowingos,zhang-etal-2020-answerfact}, consolidation is also an essential step to identify salient knowledge or evidence while mitigating duplication.


To effectively recognize multiple aspects of answers in QA systems, our \emph{first} contribution is to introduce and formalize the \emph{answer consolidation} problem.
Specifically, given a question paired with multiple answer snippets, answer consolidation first partitions the snippets into groups where each group represents a single aspect within the answer space.
Once partitioned, the final answer set is produced by returning a representative snippet from each group.
In this formulation, the answer consolidation task is a post-processing stage that takes predicted answer-mentioning snippets (in this work, sentences) from previous QA stages and produces an answer set that maximizes answer aspect coverage while minimizing answer duplication.

To foster research on the answer consolidation problem, our \emph{second} contribution is the collection of a 
new dataset, namely, \quasi (\underline{Qu}estion-\underline{A}n\underline{s}wer consol\underline{i}dation).
\quasi consists of 4,699 questions such that
each question is paired with multiple answer-mentioning sentences grouped according to different aspects.
Starting with a Quora-based question source~\cite{chen2018quora}, noting the potential for multi-aspect answers, we first retrieve 10 relevant answer sentences from the web for each question.
These sentences are then examined by three crowd-sourced workers, who exclude sentences that do not contain an answer and group the remaining ones.
Finally, individual sentence groupings from different workers are aggregated into a single partitioning. \quasi consists of 24,006 sentences and 19,676 groups, corresponding to an average number of 4.18 aspects per question and 1.22 sentences per group.

Our \emph{third} contribution is a comprehensive benchmarking for the answer consolidation problem based on \quasi.
Specifically, we consider two evaluation settings:
(1) {\em classification}, where the model predicts whether two sentences are in the same group.
(2) {\em sentence grouping}, where the model groups the answer sentences.
We evaluate a wide selection of zero-shot and supervised methods, including various SoTA sentence embedding models~\cite{reimers-gurevych-2019-sentence,simcse2021}, cross-encoders~\cite{devlin-etal-2019-bert,liu2019roberta}, and a newly proposed answer-aware cross-encoder model.
In the supervised setting, the answer-aware cross-encoder achieves the best results based on the Matthew correlation coefficient~(MCC) score of 87.8\%~(classification setting) and an adjusted mutual information~(AMI) score of 68.9\%~(sentence grouping setting).
As this performance is notably far from perfect, our findings indicate the need for future investigation on this meaningful, but challenging, task.


\section{Related Work}
\stitle{QA with multiple answers.}
Many QA datasets 
have assumed that a question has a single correct answer~\cite{joshi-etal-2017-triviaqa,kwiatkowski-etal-2019-natural}, while in real scenarios, many questions can have multiple correct answers.
Fewer datasets for QA or machine reading comprehension~(MRC) have been proposed with the consideration of multiple answers.
In extractive MRC, MASH-QA~\cite{zhu-etal-2020-question} allows one question to be answered by multiple non-consecutive text spans.
In abstractive MRC/QA, MS MARCO~\cite{Campos2016MSMA} simply treats different workers' answers as different answers.
DuReader~\cite{he-etal-2018-dureader} merges similar answers during data construction.
QReCC~\cite{anantha-etal-2021-open} allows one worker to provide multiple different answers.
Beyond these, WebQuestions~\cite{berant-etal-2013-semantic} and GooAQ~\cite{gooaq2021} include lists of diverse answers, and TREC-QA~\cite{baudivs2015modeling} uses regular expressions to capture multiple answers.
However, none of the aforementioned efforts have investigated effective consolidation of multiple answers.
In this work, we formally define and collect a dataset for the answer consolidation problem as a complement to previous work.
From another perspective, AmbigQA~\cite{min-etal-2020-ambigqa} focuses on the case where a question can be interpreted in different ways, leading to the question disambiguation task. This is fundamentally different from our work that partitions answers to the same question into different coherent subsets.
Stance detection~\cite{liu-etal-2021-multioped} is concerned with the focused problem of collecting approving/disapproving opinions for a yes-no question, 
unlike our studied problem where all multi-answer questions are considered and the answers are not limited to binary opinions.


\stitle{Answer Summarization.}
Questions with multiple answers are common in online communities.
For example, \citet{liu-etal-2008-understanding} observe that no more than 48\% of best answers on Yahoo! Answers are unique.
Many efforts~\cite{song2017summarizing,chowdhury2019cqasumm,fabbri2021multi} have been devoted to summarizing reusable answers in community QA.
Particularly, AnswerSumm~\cite{fabbri2021multi} proposes a dataset where different answers are rewritten to bullet points by humans.
While training on summarization data may enable the model to return salient and non-redundant answers, such training only works for abstractive machine readers.
A more related work is BERT-DDP~\cite{fujita-etal-2020-diverse}, which considers the problem of getting a diverse and non-redundant answer set.
They construct a dataset based on Yahoo! Chiebukuro where workers are asked to provide an answer set given a question.
However, the correct answer set is not unique when answers are equivalent.
As they treat all but the annotated answer set as wrong, both training and inference are prone to false negatives.
In this paper, we group the answers with respect to their aspects and provide a discriminant rule, such that the correct group assignment is unique.

\stitle{Diverse passage retrieval.} Many information retrieval efforts address the problem of 
retrieving diverse documents for a query~\cite{clarke2008novelty,fan2018modeling,abdool2020managing}.
In QA, \citet{jpr2021} examine answer diversity in passage retrieval and propose a self-supervised dynamic oracle training objective.
However, as passages may contain irrelevant information to the question, the retriever faces the challenge of identifying and integrating answers in passages when assessing answer diversity.
In this work, we consider a dedicated task of answer consolidation and leave the problem of identifying answers to previous QA stages.

\section{Answer Consolidation Task}
\label{sec:task_definition}
\stitle{Motivation.}
Many questions can have multiple correct answers, including questions explicitly asking for a multi-answer list~(\eg, \emph{What are the symptoms of flu?}) or questions where different people have different opinions~(\eg, debate questions), amongst others.
To provide users with a comprehensive view of the answers, the QA system needs to actively identify different answers as opposed to only returning the most popular or top-ranked answer.
Additionally, as the same answer may be repeated or paraphrased many times in the reference corpus (e.g., web), the QA system may also need to eliminate the redundant answers.
We address these requirements within answer consolidation.

\stitle{Basic concepts.}
When answering a specific question, different answers may be given regarding different perspectives, opinions, angles, or parts of the overall answer. 
We regard such answers as those pertaining to different \emph{aspects}.
Furthermore, we refer to two sentences as \emph{equivalent} if they contain the same answer aspect(s) and \emph{distinct} if they express different answer aspect(s).
To better identify equivalent/distinct sentences, we propose the following operational discriminant rule: 
%
Given two answer-mentioning sentences $s_1, s_2$ for the same question $q$, we can rewrite the answers contained in $s_1$ and $s_2$ into yes-no questions $q_1'$ and $q_2'$, which can be answered by yes/no/irrelevant.\footnote{A general process for changing a sentence to questions can be found at \url{https://www.wikihow.com/Change-a-Statement-to-Question}.
}
Then, if $s_1$ and $s_2$ give coherent answers of yes/no\footnote{Answers of irrelevant are not considered coherent.} to $q_2'$ and $q_1'$, respectively, then $s_1$ and $s_2$ are considered to represent equivalent aspects.
Otherwise, they are considered distinct from each other.

We take the following example:

\textbf{Q}: Is coffee good for your health?

\textbf{S1}: Coffee can make you slim down.

\textbf{S2}: Coffee can relieve headache.

\textbf{S3}: Coffee can help with weight loss.

Then we rewrite the answers contained in sentences as the following questions:

\textbf{Q'1}: Can coffee make you slim down?

\textbf{Q'2}: Can coffee relieve headache?

\textbf{Q'3}: Can coffee help you with weight loss?

We can tell that S1 and S3 are equivalent, as they 
both give coherent answers (yes) to each others' yes-no questions.
We can also tell that S2 is distinct from S1 and S3, as 
it gives irrelevant answers to Q'1 and Q'3.

\stitle{Task definition.}
A formal definition for answer consolidation is that 
given a question and a set of answer-mentioning sentences, answer consolidation aims at putting sentences into groups such that:
(1) each sentence belongs to exactly one group, and (2) sentences from the same/different groups are equivalent/distinct.
In this way, each sentence group corresponds to the same answer aspect(s).
We show in~\Cref{ssec:complex} that although this definition may fail for a pair of sentences if they are partially relevant, it only occurs for 2.6\% of sentences, which shows that our operational task definition works well for the majority of sentences in practice. 

In this work, we treat answer consolidation as a stand-alone process applied after QA retrieval  such that we only take the question and answer-mentioning sentences as input. In this way, the answer consolidation model is independent of the retriever and the reader architectures, and can flexibly adapt to different QA systems. We show in~\Cref{ssec:ablation} that taking sentences instead of answer spans as input leads to better performance.



\section{Question-Answer Consolidation Dataset (\quasi)}
In this section, we describe the creation of our Question-Answer consolidation dataset (\quasi) including corpus collection~(\Cref{ssec:data_collection}) and dataset annotation~(\Cref{ssec:data_annotation}). We then provide statistical and quantitative analysis of \quasi~(\Cref{ssec:complex}). 

\subsection{Corpus}
\label{ssec:data_collection}
We created \quasi based on the Quora question pairs (QQP) corpus~\cite{chen2018quora}, which consists of 364k questions pairs, originally designed for predicting whether pairs of questions have the same meaning. We start with QQP since Quora questions have a high expected propensity of having multiple correct answers.
In preprocessing, we removed questions containing spelling errors or non-English words using the Enchant library.\footnote{\url{https://abiword.github.io/enchant/}}
and questions containing personal pronouns including \{``I'', ``you'', ``we'', ``my'', and ``your''\}
as we found that such questions frequently ask about very specific personal experiences 
for which the answers may not necessarily contain any noteworthy claims, opinions, or facts in the answer.

Next, we retrieved sentences that were likely to contain multiple answers to the questions.
Given a question, we retrieved relevant sentences from the web using a SoTA industry QA retriever, where each sentence was associated with a relevance score and a URL.
To ensure the questions have multiple answers, we first removed sentences retrieved from \texttt{quora.com} and kept the questions if the relevance scores of the top three retrieved sentences were larger than specified low-confidence threshold.
Finally, we kept the top 10 sentences for each remaining question and pass to crowd workers for sentence group annotation.

\subsection{Annotation}
\label{ssec:data_annotation}
We used Amazon Mechanical Turk (AMT) to annotate \quasi. Each AMT HIT consisted of a question and 11 sentences~(including the top-10 relevance scores and one additional {\em attention-check} sentence) where the crowd workers were required to: (1) identify sentences that actually contain answers to the question and
(2) put answer-mentioning sentences into sentence groups with respect to their aspects.
The workers were allowed to skip an answer-mentioning sentence if it was hard to put it into any groups~(\eg, sentences containing more than one aspect).

\begin{figure}
    \centering
    \includegraphics[width=\linewidth]{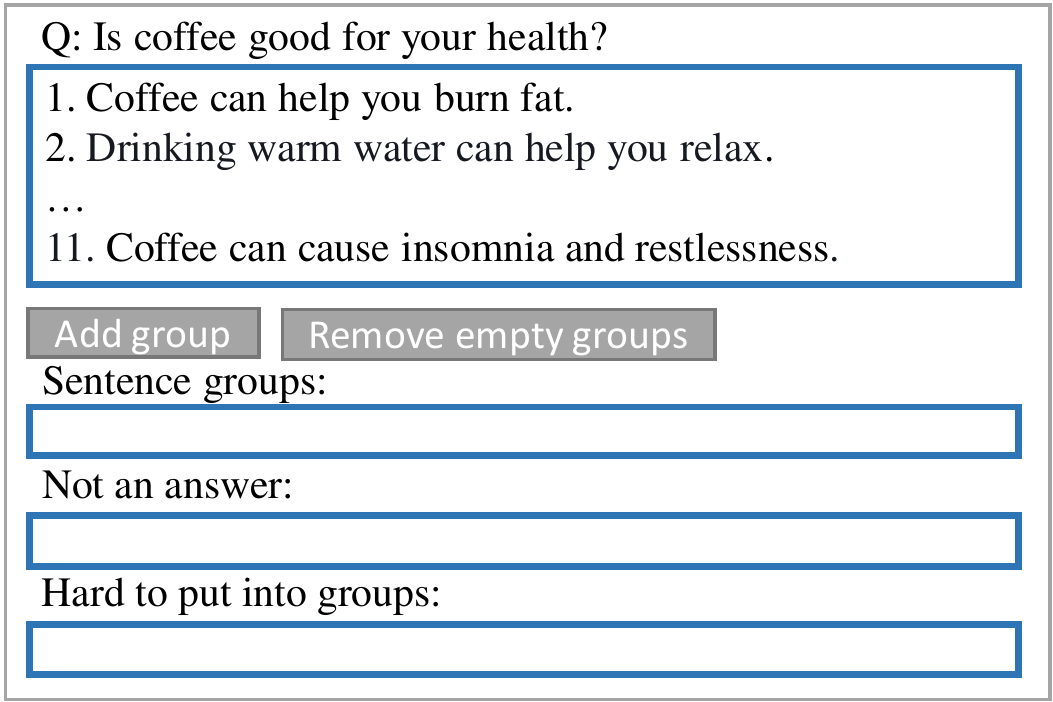}
    \caption{The interface used to collect the dataset. The second sentence is an attention-check sentence.
    }
    \label{fig:interface}
\end{figure}

The AMT interface is shown in~\Cref{fig:interface}.
Annotation was performed by dragging the sentences between blue boxes.
The sentence groups could be added or removed using the two buttons.
To submit the HIT, workers needed to put all sentences into boxes corresponding to either a specific sentence group, \emph{not an answer}, or  \emph{hard to put into groups}. 

\stitle{Cost.}
Each HIT was assigned to three annotators with pay of \$0.50/HIT, leading to target an hourly pay rate of \$15.
We randomly sampled 5k questions from~\Cref{ssec:data_collection} for HIT submission to AMT.

\stitle{Quality Control.}
We used three strategies to ensure the annotation quality:
\begin{enumerate}[leftmargin=1em]
\setlength\itemsep{0em}
    \item \textbf{Workers selection.} We only allowed crowd workers with acceptance rate $\ge 98\%$ and had completed at least 5k hits to work on the task.
    We provided annotation guidelines and examples of 3 annotated hits to instruct the workers.
    \item \textbf{Qualification test.} We manually annotated three hits as the qualification test.
    Workers were required to practice on these three hits and get the correct sentence groups on at least 2 hits to continue working on the task.
    We pay \$0.05 for each submitted hit in the qualification test.
    Although we had provided detailed instructions, only 24\% of workers passed the qualification test.
    \item \textbf{Attention checker.} For all hits, we added an attention-check sentence that was randomly sampled from other questions, so that it was unlikely to answer the question.
    As a part of the task, the worker needed to identify that this attention-check sentence did not involve an answer, otherwise she/he would be blocked from continuing to work on the task.
\end{enumerate}

\begin{table*}[!t]
    \centering
    \scalebox{0.7}{
    \begin{tabular}{p{3cm}p{4cm}p{12.5cm}c}
    \toprule
    Type& Description& Example& Percentage \\
    \midrule
    Formatting& Sentences only differ in letter case, punctuation, or short forms.& Q: How small can a black hole be? \newline
S1: scientists think the smallest black hole are \underline{as small as just one atom}.\newline
S2: Scientists think the smallest black holes are \underline{as small as just one atom}.
&9\%\\
\midrule
    Exact match& Sentences are different but answer spans are the same.& Q: What are some good stories of revenge in relationship? \newline
S1: Shakespeare’s “\underline{Hamlet}” is one of the most famous plays of revenge. \newline
S2: The play, \underline{Hamlet} by William Shakespeare explores the concept of revenge.&
47\%\\
\midrule
Lexical variation& Answer spans differ in articles, verb tenses, or have synonym substitutions.& Q: Is it hard to get a job as a fashion buyer? \newline
S1: Fashion jobs in merchandising can be \underline{very challenging}. \newline
S2: Picking one out of many fashion jobs generally is \underline{an overwhelming challenge}.&
11\%\\
\midrule
Semantic variation& Answers are paraphrased, or identification requires commonsense reasoning.& Q: How does the respiratory system work? \newline
S1: The respiratory system works by \underline{getting the good air in and} \underline{the bad air out}. \newline
S2: The Respiratory System a simple system designed to \underline{get oxygen into the body,} \underline{and to get rid of carbon dioxide and water}.
&30\%\\
\midrule
Ambiguous& We do not agree with the crowd workers' annotation.& Q: Is Zeus more powerful then Odin? (Not the same aspects.) \newline
S1: \underline{Zeus is 10 ton more than Odin}. \newline
S2: In DC, \underline{Zeus is higher than Odin}.
&3\%\\
    \bottomrule
    \end{tabular}}
    \caption{Types of equivalent sentences annotated by crowd workers. We randomly sampled 100 sentence pairs in the same group, manually annotated the answer span~(underlined), and categorized them into different types.}
    \label{tab:redundant_answers}
\end{table*}

\stitle{Label aggregation.}
To ensure data quality, we aggregated worker annotations. 
To derive the sentence set for answer consolidation, we begin by only considering sentences put into any sentence group(s) by all crowd workers as eligible, keeping 37,588 out of 50k sentences.
Next, we derived the aggregated sentence groups from AMT annotations.
As we are not aware of existing methods for this process, we proposed the following algorithm for constructing new sentence groups.
First, we sort the sentences by their relevance scores and create a sentence group with the most relevant sentence being the only member.
We then iterate over the remaining sentences with the following procedure.
For a sentence $s$, there are three possible cases:
\begin{enumerate}[leftmargin=1em]
\setlength\itemsep{0em}
    \item If there existed one group $\mathcal{G}$ such that $\forall s'\in \mathcal{G}$, $s$ and $s'$ were put into the same group by all workers, and for all already added sentences $s^* \notin \mathcal{G}$, $s$ and $s^*$ were put into different groups by all workers, we added $s$ to $\mathcal{G}$.
    \item If for all already added $s^*$, $s$ and $s^*$ were put into different groups by all workers, we created a new group with $s$ being the only member.
    \item Otherwise, we discarded sentence $s$, 
    since there was disagreement on this sentence.
\end{enumerate}
Finally, we keep each question for which the number of preserved sentences was larger than one.
Our aggregation algorithm produced sentence groups on a subset of sentences, on which all workers agree on each pair of sentences about whether they belong to the same group or not.
After this process, 4,699 out of 5,000 questions and 24,006 out of 37,588 sentences were kept.

\subsection{Dataset Analysis}
\label{ssec:complex}
We provide statistical and qualitative analyses regarding \quasi in this section.

\stitle{Annotation quality.}
We first analyze the quality of data annotation before label aggregation.
In the first annotation task of identifying whether a sentence contains an answer, AMT workers achieved an inter-annotator Fleiss’ kappa of 0.62, an average agreement rate of 90.2\%, and a worker agreement with aggregate~(WAWA) of 82.5\% in $F_1$.
WAWA is used to compare the majority vote with all workers' annotations.
In the second annotation task of sentence grouping, we first get the set of sentences that all workers put into some groups.
We then calculate the workers' agreement on each pair of sentences regarding whether they belong to the same group or not.
The inter-annotator Fleiss' kappa, average agreement rate, WAWA $F_1$ are 0.46, 84.8\%, and 75.9\%, respectively.
These results show that the overall annotation quality is usable, but with room for improvement.
Accordingly, to further improve the data quality, we only keep group annotations on which all workers agree (as stated in~\Cref{ssec:data_annotation}).

\stitle{Dataset statistics.}
Our final dataset consists of 4,699 questions, 24,006 sentences, and 19,676 groups.
On average, there are 4.18 groups per question, and 1.22 sentences per group.
Specifically, 97.7\% of questions have multiple aspects (sentence groups), and 45.4\% of questions have at least one pair of equivalent sentences.
In terms of sentence groups, 86.6\% of groups have only one sentence, 8.8\% of groups have two sentences, and the remaining 4.6\% of groups have 3 or more sentences.
Overall, this analysis shows that our dataset contains both multi-aspect and redundant answers 
that align with the challenges of answer consolidation.

\stitle{Types of equivalent sentences.}
To get a better understanding of the required knowledge to identify equivalent sentences, we randomly sampled 100 sentence pairs in the same group and manually labeled the pairs with the types shown in~\Cref{tab:redundant_answers}.
We observed that if the machine reader has the correct answer spans, 56\%~(formatting and exact match) of the equivalent sentences could be directly identified by string comparison.
Another 11\% of the equivalent sentences only differed at the lexical level, which may be identified using lemmatization, removal of stop words, or a dictionary of synonym words.
30\% of equivalent sentences are semantic variations such that identifying equivalence requires understanding of their meanings and potentially even commonsense reasoning.
E.g., for the example given in~\Cref{tab:redundant_answers}, the answer consolidation model needs to understand that \emph{oxygen} is \emph{good air} and \emph{carbon dioxide} corresponds to \emph{bad air}.
For the remaining 3\% of pairs, we do not agree with the annotation. 
Either the sentences do not answer the question, or they do not contain the same aspect(s).

\stitle{Limits of the task definition.}
During data annotation, 2.6\% of answer-mentioning sentences are denoted as ``\emph{hard to put into groups}''.
After inspection, we find that these sentences contain more than one aspect of answers.
For example, given the question \emph{What are the best places to visit and things to do in San Diego, CA?}, one sentence may be \emph{The San Diego Zoo, Balboa Park, and SeaWorld are the top tourist attractions in San Diego.}, which contains 3 different answers.
This sentence overlaps with multiple groups and thus cannot be placed in a single group.
Given the low prevalence, we leave consideration of such sentences to future work.

\section{Approach}

In this section, we first tackle the classification setting of answer consolidation~(\Cref{ssec:sentence_pair_classification}).
Given a question and answer-mentioning sentences, the task is to predict for a pair of sentences whether they are in the same group.
We consider different types of models, including sentence embedding models, cross-encoders, and answer-aware cross-encoders.

Then we consider the sentence grouping setting, presenting the method of transforming pairwise predictions to sentence groups~(\Cref{ssec:grouping}).
For all methods, we use RoBERTa$_\textsc{large}$~\cite{liu2019roberta} as the encoder, noting that other pretrained language models~(PLMs) can easily be incorporated as part of these methods.


\subsection{Sentence Pair Classification}
\label{ssec:sentence_pair_classification}
\stitle{Sentence embedding models.}
Sentence embedding models~\cite{reimers-gurevych-2019-sentence,simcse2021} produce for each sentence an embedding vector, with which we can use metrics such as cosine to calculate their similarity.
Specifically, given a question $q$ and a sentence $s$, we first tokenize them to $X_q$ and $X_s$ using the RoBERTa tokenizer, and then concatenate them as inputs:
\begin{tcolorbox}
\small
        \texttt{<s>}$X_q$ $X_s$\texttt{</s>}
\end{tcolorbox}
Following~\citet{simcse2021}, we take the \texttt{<s>} embedding in the last layer of PLM as the sentence embedding.
Then for a pair of sentences, whether they are in the same group is decided by the cosine similarity of the sentence embedding. 
The similarity can be converted to binary predictions using the best threshold that is selected on the validation set.

The sentence embedding models can work in both zero-shot and supervised settings.
In the zero-shot setting, we directly use the pretrained sentence embedding model to make predictions without fine-tuning.
In the supervised setting, given a pair of sentence embedding $\bm{h}_1$, $\bm{h}_2$, and label $y\in \left\{0, 1\right\}$, where $0$ and $1$ mean not in/in the same group respectively,\footnote{We find that using 0 for \emph{not in the same group} achieves better results than using -1.}
we fine-tune the PLM based on the following
regression objective of sentence-transformers~\cite{reimers-gurevych-2019-sentence}:
\begin{align*}
    \mathcal{L}_\text{reg} = \left(\cos\left(\bm{h}_1, \bm{h}_2\right) - y\right)^2.
\end{align*}


\stitle{Cross-encoders.}
Cross-encoders~\cite{devlin-etal-2019-bert,liu2019roberta} take a pair of sentences as the input and predict whether they are in the same group or not.
Given a question $q$ and two answer-mentioning sentences $s_1$ and $s_2$, we first tokenize them as $X_q$, $X_{s_1}$, and $X_{s_2}$ using the RoBERTa tokenizer, and then take $X_q$ $X_{s_1}$ and $X_q$ $X_{s_2}$ as two segments of inputs, following the input formats of sentence pair classification tasks~\cite{liu2019roberta}:

\begin{tcolorbox}
\small
        \texttt{<s>}$X_q$ $X_{s_1}$\texttt{</s>}\texttt{</s>}$X_q$ $X_{s_2}$\texttt{</s>}
\end{tcolorbox}

Prediction is independently performed on sentence pairs using a binary classifier on the first special (classification) token \texttt{<s>} embedding in the last layer of the PLM.
The cross-encoders work in both zero-shot and supervised settings.
In the zero-shot setting, we fine-tune the model on the MNLI~\cite{williams-etal-2018-broad} dataset and take \emph{entailment} as \emph{in the same group}.
In the supervised setting, given the sentence pair embedding (obtained from \texttt{<s>}) $\bm{h}$ and the label $y$, we fine-tune the model using the binary cross-entropy loss:
\begin{align*}
p &= \sigma\left(\bm{w}^\intercal\bm{h}\right), \\
    \mathcal{L}_\text{bce} &= -\left(y \log p + (1-y) \log\left(1-p\right)\right),
\end{align*}
where $p$ is the probability that the sentences are in the same group, $\sigma$ is the sigmoid function, $\bm{w}$ is a parameter of the classifier.
In inference, we convert $p$ to binary predictions using the best threshold selected on the validation set.
The cross-encoders require predicting on all sentence pairs and have higher computational costs than sentence embedding models.
However, we observe in experiments that cross-encoders consistently outperform sentence embedding models in the supervised setting.

\stitle{Answer-aware~(A$^2$) cross-encoders.}
\Cref{ssec:complex} shows that 56\% of equivalent sentences can be directly identified if the model knows the answer spans.
Therefore, we provide the answer consolidation models with answers generated from the UnifiedQA$_\textsc{large}$ model~\cite{khashabi-etal-2020-unifiedqa}.
As the UnifiedQA is trained on both extractive and abstractive datasets, the answers may not be text spans of sentences.
Specifically, given a question $q$, two sentences $s_1$ and $s_2$, and the generated answers $a_1$ and $a_2$, we first tokenize them as $X_p$, $X_{s_1}$, $X_{s_2}$, $X_{a_1}$, and $X_{a_2}$, respectively, then construct the input to cross-encoders as:
\begin{tcolorbox}
\small
\texttt{<s>}$X_q$ $X_{s_1}$ $X_{a_1}$\texttt{</s>}\texttt{</s>}$X_q$ $X_{s_2}$ $X_{a_2}$\texttt{</s>}
\end{tcolorbox}
\noindent
The training process and inference process are the same as the cross-encoders.

\subsection{Sentence Grouping}
\label{ssec:grouping}
In answer consolidation, our ultimate goal is to obtain the consolidated sentence groups.
This is done in a two-step approach.
The first step is to get the matrix of distances $\mathcal{D}$ between pairs of sentences for a question. 
We perform this step using models trained in the classification setting.
For sentence embeddings, $\mathcal{D}$ is adopted as the pairwise cosine distance matrix.
For cross-encoders, each entry of $\mathcal{D}$ equals to 1 minus the predicted probability for a sentence pair.
As $\mathcal{D}$ derived in this way may not be always symmetric, we use $\frac{1}{2}\left(\mathcal{D} + \mathcal{D}^\intercal\right)$ as the distance matrix instead.

The next step is to transform the distance matrix into sentence groups.
Here we apply agglomerate clustering~\cite{han2011data}.
It uses a bottom-up strategy, starting from letting each sentence form its own cluster, and then recursively merging the clusters if their distance is smaller than a threshold.
We use the average distance of sentence pairs as the inter-cluster distance measure and select the best threshold on the validation set.
Agglomerate clustering stops when the distances between all clusters are larger than the threshold.

\section{Experiments}

In this section, we present the experimental setup~(\Cref{ssec:setup}), show the main results~(\Cref{ssec:main_results}), conduct an ablation study~(\Cref{ssec:ablation}), and provide error analysis~(\Cref{ssec:case_study}).

\subsection{Experimental Setup}
\label{ssec:setup}
\stitle{Dataset.}
We randomly split the 4,699 questions into an 80/10/10 split, which serves as the training, validation, and test set, respectively.

\stitle{Evaluation metrics.}
We use different evaluation metrics in the two evaluation settings.
For the classification setting, we first use the micro $F_1$ measure. 
Considering that classes in the dataset are highly imbalanced~(only $11\%$ of sentence pairs are in the same group), we additionally use the Matthews correlation coefficient~(MCC; \citealt{matthews1975comparison}), which is considered a more class-balanced metric.
For the sentence grouping setting, we use clustering metrics including adjusted rand index~(ARI; \citealt{rand1971objective}) and adjusted mutual information~(AMI; \citealt{nguyen2009information}).
These two metrics take the predicted grouping and the ground-truth grouping, and measure the similarity between them.
For all metrics, larger values indicate better performance, and a value of $100\%$ indicates perfect classification/grouping.

\stitle{Configuration.}
We implement the models using Huggingface's Transformers~\cite{wolf-etal-2020-transformers}.
The models are optimized with Adam~\cite{Kingma2015AdamAM} using a learning rate of $1\mathrm{e}{-5}$, with a linear decay to 0.
We fine-tune all models for 10 epochs with a batch size of 32 questions~(including all associated sentence pairs).
The best model checkpoint and thresholds are selected based on the validation set.
We report the average results on 5 runs of training using different random seeds.

\stitle{Models.}
We use RoBERTa$_\textsc{large}$~\cite{liu2019roberta} as the encoder for all models.
For sentence embedding models, we try RoBERTa fine-tuned on two intermediate tasks:
1) SRoBERTa~\cite{reimers-gurevych-2019-sentence} is fine-tuned on natural language inference~(NLI) datasets, achieving better results on semantic textual similarity~(STS) tasks, and 2) SimCSE-RoBERTa~\cite{simcse2021} is fine-tuned in a self-supervised fashion, taking a sentence and predicting itself using a contrastive learning objective.
For cross-encoders, in addition to directly running supervised fine-tuning on our data, we also try supplementary training on an intermediate labeled-data task~\cite{Phang2018SentenceEO}, which fine-tunes cross-encoders on MNLI before supervised fine-tuning.
Particularly in the latter setting, we observe it being necessary to re-initialize the classifier before supervised fine-tuning to obtain more promising performance.

\begin{table}[!t]
\centering
\scalebox{0.78}{
    \begin{tabular}{p{4.3cm}cccc}
         \toprule
         \textbf{Model} & \multicolumn{2}{c}{\textbf{Classification}} & \multicolumn{2}{c}{\textbf{Grouping}} \\
          &$F_1$ & MCC & ARI& AMI \\
         \midrule
         \multicolumn{5}{c}{\emph{Zero-shot Sentence Embedding}} \\
         RoBERTa& 22.2& 9.2 & 55.8& 33.3 \\
         SRoBERTa& 41.9& 35.1& 57.8& 37.7 \\
         SimCSE-RoBERTa& \textbf{53.2}& \textbf{47.6}& 66.1& 46.1 \\
         \midrule
         \multicolumn{5}{c}{\emph{Zero-shot Cross-Encoders}} \\
         RoBERTa-MNLI& 52.4& 47.0& \textbf{69.0}& \textbf{48.5} \\
         A$^2$RoBERTa-MNLI& 40.4& 34.6& 60.9& 40.0 \\
        \midrule
        \midrule
        \multicolumn{5}{c}{\emph{Supervised Sentence Embedding}} \\
         RoBERTa& 73.5& 70.3& 79.2& 58.5 \\
         SRoBERTa& 80.7& 78.8& 85.3& 64.3 \\
         SimCSE-RoBERTa& 81.1& 79.0& 85.7& 64.9 \\
         \midrule
        \multicolumn{5}{c}{\emph{Supervised Cross-Encoders}} \\
         RoBERTa& 86.8& 85.9& 88.4& 66.9 \\
         A$^2$RoBERTa& 88.2& 86.8& 89.6& 68.2\\
         RoBERTa-MNLI& 88.7& 87.4& 89.6& 68.2 \\
         A$^2$RoBERTa-MNLI& \textbf{89.0}& \textbf{87.8}& \textbf{90.4}& \textbf{68.9} \\
         \bottomrule
    \end{tabular}}
    \caption{Main results on the test set of \quasi. The best results in the zero-shot and supervised settings are highlighted in \textbf{bold}.
    }
    \label{tab::main_results}
\end{table}

\subsection{Main Results}\label{ssec:main_results}
The experimental results on both the pairwise classification and sentence grouping settings are reported in~\Cref{tab::main_results}.
We observe that in the zero-shot setting, intermediate-task training improves answer consolidation, while the performance remains far behind supervised models.
In the supervised setting, cross-encoders consistently outperform the sentence embedding models.
Overall, the answer-aware cross-encoder intermediately tuned on MNLI (A$^2$RoBERTa-MNLI) achieves the best results on all metrics, showing that intermediate-task training on MNLI improves performance.
Besides, we find that answer-aware cross-encoders outperforms regular cross-encoders, showing that answers generated by the machine reader provide additional information 
that helps consolidation.

\subsection{Ablation Study}
\label{ssec:ablation}
In this section, we study the model performance based on different input information given to supervised models.
We denote the questions, sentences as Q, S, and answers generated by UnifiedQA as A.
The results are shown in~\Cref{tab::ablation}.
Overall, models trained on all inputs~(Q+S+A) achieve better results than those that have observed only a subset of the available inputs on most metrics.
Removing the sentences leads to the largest drops in performance (\eg, 20.9\% in $F_1$ for SimCSE-RoBERTa and 22.3\% in $F_1$ for RoBERTa-MNLI), which shows that sentences provide useful information for answer consolidation.
Using sentences only leads to the second-largest drop in performance, showing that without grounding to questions and answers, consolidation is not simply addressed only with the sentences.
Besides, removing questions also leads to more significant drops in performance than removing answers (\eg, 4.9\% in $F_1$ for SimCSE-RoBERTa and 2.3\% in $F_1$ for RoBERTa-MNLI).
This shows that it is necessary 
to understand the answer equivalence within the question context in order to consolidate answers.

\begin{table}[!t]
\centering
\scalebox{0.72}{
    \begin{tabular}{p{5cm}cccc}
         \toprule
         \textbf{Model} & \multicolumn{2}{c}{\textbf{Classification}} & \multicolumn{2}{c}{\textbf{Grouping}} \\
          &$F_1$ & MCC & ARI& AMI \\
         \midrule
        \multicolumn{5}{c}{\emph{Supervised Sentence Embedding}} \\
         SimCSE-RoBERTa~(Q+A)& 61.6& 57.6& 63.4& 51.6\\
         SimCSE-RoBERTa~(S)& 72.1& 69.0& 78.2& 58.1\\
         SimCSE-RoBERTa~(S+A)& 77.6& 75.0& 80.4& 60.0\\
         SimCSE-RoBERTa~(Q+S)& 81.1& 79.0& \textbf{85.2}& \textbf{64.9}\\
         SimCSE-RoBERTa~(Q+S+A)& \textbf{82.5}& \textbf{80.4}& 85.1& 64.6\\
         \midrule
        \multicolumn{5}{c}{\emph{Supervised Cross-Encoders}} \\
         RoBERTa-MNLI~(Q+A)& 66.7& 62.9& 75.1& 53.4\\
         RoBERTa-MNLI~(S)& 83.8& 81.9& 85.3& 65.0 \\
         RoBERTa-MNLI~(S+A)& 85.7& 84.1& 87.5& 66.7 \\
         RoBERTa-MNLI~(Q+S)& 88.7& 87.4& 89.6& 68.2 \\
         RoBERTa-MNLI~(Q+S+A)& \textbf{89.0}& \textbf{87.8}& \textbf{90.4}& \textbf{68.9} \\
         \bottomrule
    \end{tabular}}
    \caption{Results with different input formats on the test set. Q, S, A denotes question, sentences, and answers (generated by UnifiedQA), respectively. RoBERTa-MNLI~(Q+S+A) is equivalent to A$^2$RoBERTa-MNLI.}
    \label{tab::ablation}
\end{table}

\begin{table*}[t]
    \centering
    \scalebox{0.8}{
    \begin{tabular}{p{3cm}p{3cm}p{12.5cm}}
    \toprule
    Cause& Description& Example \\
    \midrule
Entailment~(16.7\%)& One answer is entailed by the other.& Q: What makes successful people different from average people? \newline
S1: Wealthy people are not afraid of failure, unlike average people who often do not even try. \newline
S2: The difference between average people and achieving people is their perception of and response to failure.
\\
\midrule
    Entity/commonsense knowledge~(23.8\%)& The answers refer to the same entity or are equivalent by common sense.& Q: What are some best Hollywood romantic movies to watch? \newline
S1: When the subject of romantic movies comes up, one of the first that comes to mind in any list of all-time greats is Casablanca. \newline
S2: Humphrey Bogart and Ingrid Bergman’s film about love and loss during WWII is basically required viewing for anyone who enjoys romantic movies.\\
\midrule
Semantic equivalence~(57.1\%)& The answers have the same semantic meaning but are expressed using different words.& Q: What is the equity risk premium? \newline
S1: Equity Risk Premium is the difference between returns on equity/individual stock and the risk-free rate of return. \newline
S2: Let us start with defining the equity risk premium: the Equity Risk Premium is the average extra return demanded by investors, on top of a risk free rate, as a compensation for investing in equity securities with average risk. \\
\midrule
Spelling errors (2.4\%)& There are spelling errors in the answers. &Q: Are all psychopaths narcissists? \newline
S1: I came across that all psychopats are narcissists, but not all narcissists are psychopats. \newline
S2: I have read it summed up this way: Not all narcissists are psychopaths, but all psychopaths are narcissists.
\\
    \bottomrule
    \end{tabular}}
    \caption{Different causes of wrongly classified positive pairs.
    }
    \label{tab::error_analysis}
\end{table*}

\subsection{Error Analysis}
\label{ssec:case_study}
To get a sense of what knowledge is needed to further improve model performance, we examined sentence pairs incorrectly classified as \emph{not in the same group} by A$^2$RoBERTa in the validation and test sets, where 318 out of 2,614 pairs~(12.2\%) are wrongly classified.
We randomly sample 50 such error cases and categorize them by the answer-equivalence type as defined in~\Cref{tab:redundant_answers}.
Of the 50 pairs, 1~(2\%) is from \emph{exact match}, 8~(16\%) are ambiguous, and the remaining 41~(82\%) are from the \emph{semantic variation} category, showing that it is the most challenging type to tackle.

We further study the specific causes of errors on the 42 unambiguous pairs.
Examples of distinct error causes are described in~\Cref{tab::error_analysis}
We find that $16.7\%$ of the falsely classified sentence pairs contain one answer that entails the other instead of expressing the exact same answers, which should however be considered redundant answers by our definition.
$80.9\%$ of pairs are equivalent but require understanding the semantic meanings or entity-specific/commonsense knowledge.
The rest $2.4\%$ contain spelling errors that negatively affect model inference.


\section{Conclusion}
In this paper, we formulate and propose the answer consolidation task that seeks to group answers into different aspects.
This process can be used to construct a final set of answers that is both comprehensive and non-redundant.
We contribute the Question-Answer consolidation dataset (\quasi) for this task and evaluate various models, including sentence embedding models, cross-encoders, and answer-aware cross-encoders.
While the best-performing supervised models have  achieved promising performance, 
without that abundant annotation, unsupervised methods still remain far from perfect.
This suggests room for further studies on more robust and generalizable solutions for answer consolidation that would largely benefit real-world open-domain QA systems.

\section*{Acknowledgement}

We appreciate the reviewers for their insightful comments and suggestions.
Wenxuan Zhou and Muhao Chen are supported by the National Science Foundation of United States Grant IIS 2105329.

\bibliography{anthology,custom}
\bibliographystyle{acl_natbib}

\end{document}